\documentclass[11pt, a4paper, logo, copyright]{eai}
\usepackage{shlab}

\usepackage[authoryear, sort&compress, round]{natbib}
\usepackage[]{mdframed}

\usepackage{dblfloatfix}

\usepackage{amsmath,amsfonts,bm}
\usepackage{multirow}
\usepackage{subcaption}
\usepackage{wrapfig}

\def\eqref#1{equation~\ref{#1}}

\def\1{\bm{1}}

\DeclareMathAlphabet{\mathsfit}{\encodingdefault}{\sfdefault}{m}{sl}
\SetMathAlphabet{\mathsfit}{bold}{\encodingdefault}{\sfdefault}{bx}{n}

\newcommand{\modelname}{InternVLA-A1\xspace}

\usepackage{listings}
\usepackage{hyperref}
\usepackage{url}
\usepackage{graphicx}
\usepackage{amsmath} %
\usepackage{mathrsfs} %
\usepackage{etoolbox}
\usepackage{cleveref}
\usepackage{tcolorbox}
\usepackage{colortbl}
\usepackage{booktabs}       %
\usepackage{amsfonts}       %
\usepackage{nicefrac}       %
\usepackage{microtype}      %
\usepackage{caption}
\usepackage{xspace} % 
\usepackage{subcaption}
\usepackage{algorithm}
\usepackage{algorithmic}
\usepackage{multirow}
\usepackage{subcaption}
\usepackage{lipsum}

\usepackage{adjustbox}

\usepackage{booktabs} %
\usepackage{enumitem}%
\setlist[itemize]{noitemsep, topsep=0pt}
\usepackage{enumitem,kantlipsum}

\usepackage{multicol,multirow} 
\usepackage{graphicx}    
\usepackage{pifont}
\usepackage{minted}
\usepackage{ulem}
\usepackage{xcolor}
\usepackage{longtable}
\usepackage{makecell} % for multirowcell
\usepackage{listings}
\usepackage{xcolor}
\usepackage[T1]{fontenc}

\lstdefinestyle{yamlstyle}{
    basicstyle=\ttfamily\footnotesize,
    numbers=left,
    numberstyle=\tiny,
    stepnumber=1,
    numbersep=6pt,
    frame=single,
    framerule=0.5pt,
    breaklines=true,
    breakatwhitespace=true,
    tabsize=2,
    captionpos=b,
    keywordstyle=\color{blue},
    commentstyle=\color{gray},
    stringstyle=\color{teal},
}

\newlength\savewidth

\definecolor{baselinecolor}{HTML}{d6eaf8}

\definecolor{mygray}{gray}{0.4}

\AtBeginEnvironment{tcolorbox}{\tiny}

\newcount\Comments  %
\usepackage{color}
\definecolor{darkred}{rgb}{0.9,0,0}
\definecolor{darkgreen}{rgb}{0,0.5,0}
\definecolor{darkblue}{rgb}{0,0,0.7}
\definecolor{purple}{rgb}{.6, 0,.6}
\definecolor{orange}{rgb}{1.0,0.64,0}

\definecolor{deemph}{gray}{0.6}

\definecolor{baselinecolor}{gray}{.9}
\definecolor{yellow}{RGB}{218,165,32}
\definecolor{lightcyan}{rgb}{0.88, 1.0, 1.0}
\definecolor{lightskyblue}{rgb}{0.53, 0.81, 0.98}
\definecolor{aliceblue}{rgb}{0.94, 0.97, 1.0}
\definecolor{LightSlateBlue}{RGB}{70,130,180}
\definecolor{DeepBlue}{RGB}{65,100,170}
\definecolor{DeepPurple}{RGB}{136,105,160}
\definecolor{LightGreen}{RGB}{59,125,35}
\definecolor{LightRed}{RGB}{234,66,53}
\definecolor{cvprblue}{rgb}{0.21,0.49,0.74}

\definecolor{mypink}{RGB}{254,102,140}
\definecolor{myclay}{RGB}{59,182,176}

\newcommand{\kibitz}[2]{\ifnum\Comments=1\textcolor{#1}{#2}\fi}

\title{InternVLA-A1: Unifying Understanding, Generation and Action for Robotic Manipulation}

\renewcommand{\today}{}
\author[]{InternVLA-A1 Team \\
Full author list in \hyperref[sec:contributors]{Contributors} section}

\renewcommand{\footnoterule}{%
  \kern -3pt
  \hrule width 1.0\columnwidth height 0.5pt
  \kern 2.6pt
}

\begin{document}

\begin{abstract}
\vspace{-24pt}
Prevalent Vision-Language-Action (VLA) models are typically built upon Multimodal Large Language Models (MLLMs) and demonstrate exceptional proficiency in semantic understanding, but they inherently lack the capability to deduce physical world dynamics. Consequently, recent approaches have shifted toward World Models, typically formulated via video prediction; however, these methods often suffer from a lack of semantic grounding and exhibit brittleness in the presence of video prediction errors. 
To synergize semantic understanding with dynamic predictive capabilities, we present InternVLA-A1. This model employs a unified Mixture-of-Transformers architecture, coordinating three experts for scene understanding, visual foresight generation, and action execution. These components interact seamlessly through a unified masked self-attention mechanism. 
Building upon InternVL3 and Qwen3-VL, we instantiate InternVLA-A1 at 2B and 3B parameter scales. We pre-train these models on heterogeneous data sources over real-world robot data, synthetic simulation data, and human videos, covering over 692M frames. This hybrid training strategy effectively harnesses the diversity of synthetic simulation data while minimizing the sim-to-real gap. 
We evaluated InternVLA-A1 on 12 real-world robotic tasks and a simulation benchmark. The results show that InternVLA-A1 consistently outperforms prior leading models: compared with $\pi_{0.5}$, it achieves +4.4\% on static manipulation tasks and +2.6\% on the RoboTwin 2.0 simulation benchmark, and delivers a +26.7\% boost on dynamic manipulation tasks.

\links{
  \link{homepage}{Homepage}{https://internrobotics.github.io/internvla-a1.github.io/}, 
  \link{code}{Code:InternVLA-A1}{https://github.com/InternRobotics/InternVLA-A1}, 
  \link{huggingface}{Model:InternVLA-A1}{https://huggingface.co/InternRobotics/InternVLA-A1-3B}, 
  \link{data}{Data:InternData-A1}{https://huggingface.co/datasets/InternRobotics/InternData-A1},
}

\end{abstract}

\maketitle

\vspace{-12pt}

\begin{figure}[h]
    \centering
    \includegraphics[width=0.87\linewidth]{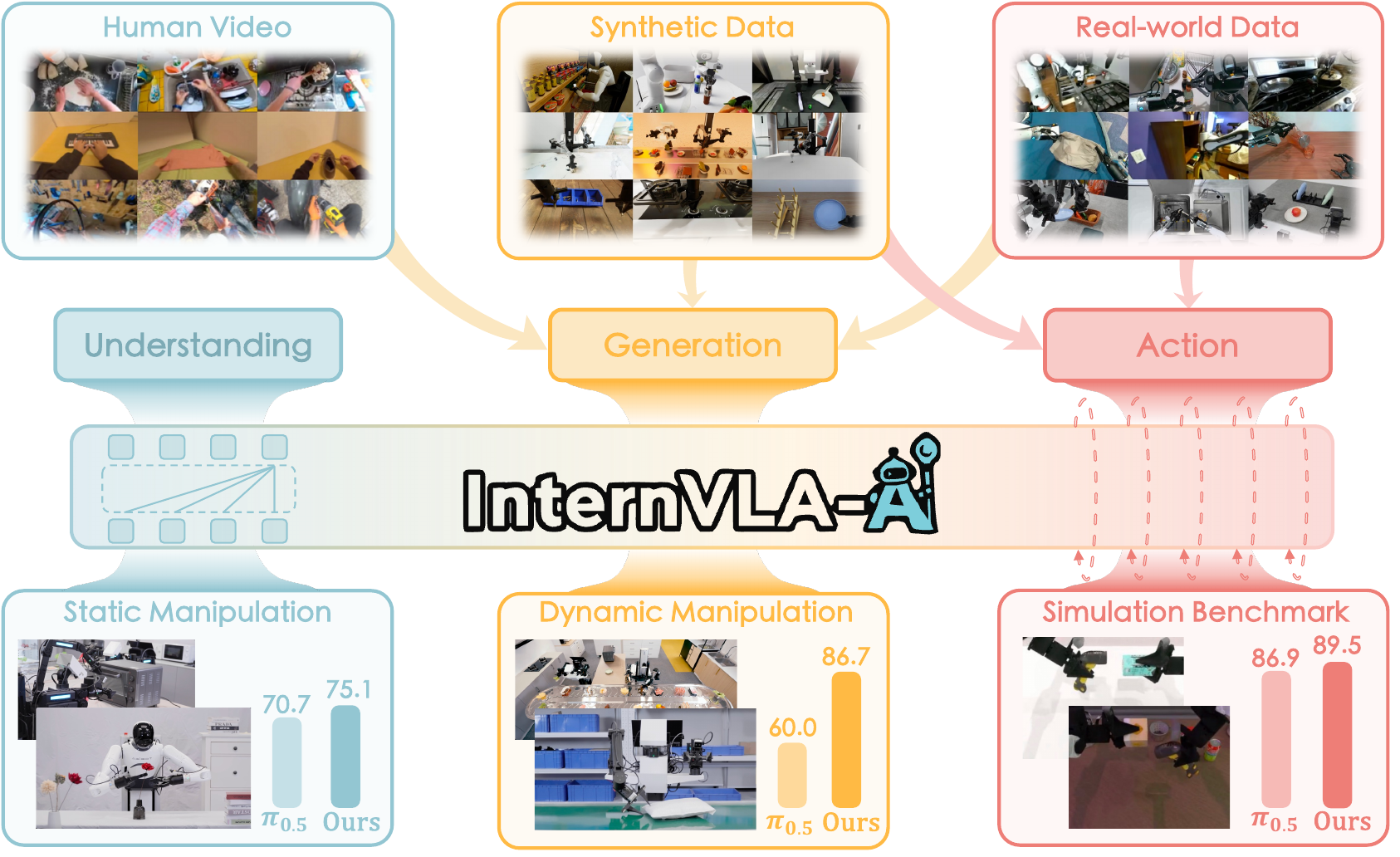}
    \caption{\textbf{InternVLA-A1} unifies scene understanding, visual foresight generation, and action execution into a single framework. This architecture couples semantic reasoning with dynamics prediction to guide action execution, and effectively enables joint training on heterogeneous data sources over human videos, synthetic data, and real-world demonstrations. The resulting model exhibits consistent robustness across static manipulation, dynamic manipulation, and simulation benchmarks, especially demonstrating remarkable superiority in dynamic scenarios.} %
    \label{fig:teaser}
\end{figure}

\section{Introduction}

The pursuit of intelligent generalist robots remains a cornerstone of robotics research~\citep{bu2024robodual,clark2025rad,huang2023voxposer,cui2025openhelix,fang2023anygrasp,qu2025eo1,huang2025otter,li2025hamster}.
Currently, the community favors the end-to-end learning paradigm and the Vision-Language-Action (VLA) architecture~\citep{chen2025internvlam1,black2024pi_0,pi_05,bjorck2025gr00t_n1,grootn1.5,li2025cronusvla,yang2025instructvla,zhao2025cotvla,yang2025gripper,cen2025worldvla,cheang2025gr3,lin2025onetwovla,zheng2025xvla,zhai2025walloss} to realize such generalist policies.
Built upon Multimodal Large Language Models (MLLMs) and trained on massive real-world robot demonstrations, VLA models exhibit remarkable performance in daily tasks such as clothes folding and table bussing. 
Nevertheless, the generalizability of these models still falls short of practical application requirements. Their adaptability to scene variations remains inadequate, such as dynamic settings involving industrial conveyor belts.

The generalization bottlenecks in existing policies stem from two primary issues: deficient physical world cognition and a lack of adaptive manipulation capabilities. Addressing the first challenge requires integrating foundation models, such as Multimodal Large Language Models (MLLMs)~\citep{alayrac2022flamingo,beyer2024paligemma,chen2025eagle25,touvron2023llama2} or world models~\citep{zheng2024opensora,blattmann2023svd,hacohen2024ltx,assran2025vjepa2}, to enhance cognitive capacity, while addressing the latter necessitates large-scale and diverse robot action datasets for skill learning. Consequently, achieving generalist policies requires a synergistic strategy that advances both model architecture and training data.

Regarding \textit{model architecture}, prevalent generalist policies, such as $\pi_0$~\citep{black2024pi_0}, $\pi_{0.5}$~\citep{pi_05}, and GR00T N1/N1.5~\citep{bjorck2025gr00t_n1}, are built upon MLLMs that map visual data into a text-based feature space. Although this grants them strong semantic understanding, text tokens are ill-suited for modeling physical laws, resulting in a deficiency in physical dynamics reasoning.
Consequently, these policies are optimized for reactive perception-to-action mapping, rather than reasoning about how states will evolve under motion and contact. 
This limitation becomes particularly evident in dynamic environments, such as industrial conveyor settings, where understanding momentum, inertia, and contact dynamics is critical.
Recent efforts attempt to incorporate foresight via World Models, notably through video prediction paradigms like VPP~\citep{hu2024vpp} and Genie Envisioner~\citep{liao2025genie_envisioner}. These methods generate anticipated observations to guide decision-making, yet they often suffer from weak semantic grounding and are sensitive to video prediction errors.
Consequently, developing a unified architecture that tightly couples semantic understanding with robust predictive dynamics is crucial for reliable dynamics-aware manipulation.

Regarding \textit{training data}, existing VLA models have gained adaptability by scaling up large real-robot datasets, and current generalist policies rely heavily on such real-robot data~\citep{walke2023bridgedata, bu2025agibotworld, wu2024robomind}. For example, pioneering works collected 130K episodes to train RT-1~\citep{brohan2022rt1} and RT-2~\citep{zitkovich2023rt2}, and subsequent initiatives aggregated over one million demonstrations from 22 heterogeneous robots to build Open X-Embodiment~\citep{Neill2024oxe}. 
However, relying solely on real-world data collection remains challenging. 
Although $\pi_0$ was trained with 10{,}000 hours demonstrations, spanning 68 tasks from seven robot morphologies and achieving strong dexterous manipulation, it still struggles to adapt to scene variations, especially in highly dynamic environments. 
Further expanding real-robot datasets and covering long-tail scene variations at scale is costly and inefficient.
In contrast, simulation presents a promising complementary approach. Its extensive libraries of scenes and objects enrich sample diversity, and domain randomization simulates scene variations to improve the policy's robustness in changing environments. Additionally, simulation ensures more controllable trajectories by eliminating noisy data.
Our prior work, InternData-A1~\citep{tian2025interndata_a1}, validated that large-scale and high-fidelity simulation data can effectively support the pre-training of VLA models.
Nevertheless, simulation data suffers from the inevitable sim-to-real gap, especially in contact-rich dynamics. Therefore, synergizing the diversity of simulation with the physical fidelity of real-world data presents a promising avenue to overcome these respective limitations.

To address the above generalization challenges, particularly robustness against dynamic scene variations, we propose InternVLA-A1. As depicted in \Cref{fig:teaser}, our model features a novel architecture integrating understanding, generation, and action.
InternVLA-A1 combines the semantic reasoning of MLLMs with the prediction capability of a World Model-style imagination module, effectively bridging the gap between semantics and physical dynamics to facilitate foresight-aware action generation.
Furthermore, we train InternVLA-A1 on three complementary data sources: large-scale simulated robot trajectories, real-robot demonstrations, and egocentric human videos. This joint training recipe enables scalable diversity from simulation, grounds action execution with real-robot interactions to reduce the sim-to-real gap, and enriches visual representations from human videos to better capture complex manipulations. 
By combining these three sources in a unified training pipeline, InternVLA-A1 benefits from scalable variation without sacrificing physical fidelity, leading to improved robustness and real-world transfer.
We validate InternVLA-A1 through extensive experiments on 12 real-world tasks and a simulation benchmarks, and observe consistent improvements over strong VLA baselines. 
Specifically, InternVLA-A1 surpasses $\pi_{0.5}$ with a +4.4\% improvement on real-world static tasks and +2.6\% gain on the RoboTwin 2.0 simulation benchmark, alongside a substantial +26.7\% boost on real-world dynamic tasks.

\section{Related Works}
\label{sec:related_works}

% From the very outset of \textbf{embodied AI}, a core challenge has been to develop a paradigm for exploring the scaling law of embodied intelligence at low cost and high efficiency.

In this section, we compare InternVLA-A1 with current methods from the perspectives of model architecture and training data.
%In the \textbf{vision-language-action model}, RT-2~\citep{zitkovich2023rt2} co-fine-tunes state-of-the-art vision-language models on both robotic trajectory data and Internet-scale vision-language tasks, enabling emergent capabilities like improved generalization, command interpretation, and multi-stage reasoning in robotic manipulation.
%Unlike RT-2, which trains from scratch, $\pi_0$~\citep{black2024pi_0} leverages a pre-trained vision-language model (VLM) to inherit Internet-scale knowledge, enabling it to perform complex tasks across diverse robot platforms, acquire new skills through fine-tuning, and follow language instructions from both humans and high-level VLM policies. To better integrate the capabilities of VLM and real-time fluid motor control, GR00T N1~\citep{bjorck2025gr00t_n1} proposes a VLA model with a dual-system architecture, combining a vision-language module for environmental interpretation and reasoning, and a diffusion policy for real-time motor action generation, enabling high performance in language-conditioned tasks. Building on this foundation, GR00T N1.5 improves upon GR00T N1 with enhancements in model architecture, data integration, and performance, including better VLM grounding, improved language command following, higher data efficiency, and expanded compatibility with various robot embodiments. 

In \textbf{Vision-Language-Action models}, a common practice is to integrate the multimodal capabilities of foundation models with robotic control.
%
% RT-2~\citep{zitkovich2023rt2} and OpenVLA~\citep{kim2024openvla},  pioneered the co-training of MLLMs on both robotic trajectories and Internet-scale data, enabling emergent reasoning capabilities. 
%
RT-2~\citep{zitkovich2023rt2} and OpenVLA~\citep{kim2024openvla} use a vocabulary-replacement technique that maps text tokens in an LLM to discrete action representations, 
thereby leveraging the LLM’s general-purpose capabilities to facilitate emergent reasoning. 
Diverging from discrete token-based approaches, 
$\pi_0$~\citep{black2024pi_0} adopts a Mixture-of-Transformers architecture that combines a pre-trained MLLM with an action expert, and uses flow matching to output continuous actions, enabling more precise and dexterous control.
Building on this, $\pi_{0.5}$~\citep{pi_05} combines high-level sub-task prediction with low-level action prediction, thereby enhancing its capability for long-horizon tasks and cross-scene generalization.
%
%Building on this, $\pi_{0.5}$~\citep{pi_05} uses co-training on heterogeneous tasks including robot manipulation, high-level subtask prediction, and general multimodal grounding and reasoning, thereby enable broad generalization across environments and objects. 
%
In addition, GR00T N1/N1.5~\citep{bjorck2025gr00t_n1} adopts an approach that is closer to a dual-system paradigm, pairing a VLM for high-level reasoning with a DiT for action generation.
Distinct from these approaches, InternVLA-A1 unifies the semantic understanding of MLLMs with the dynamics‑prediction capabilities, effectively bridging the semantics-dynamics gap prevalent in existing VLA architectures.

In \textbf{Video prediction and world models}, extensive research leverages future video prediction to facilitate robotic control~\citep{zhang2025dreamvla,hung2025nora15,zhu2025wmpo}. 
UniPi~\citep{Unipi} trains a text-conditioned video generator paired with an inverse dynamics model to derive actions, whereas UniSim~\citep{UniSim} employs generative modeling to construct a universal simulator for training both high-level and low-level policies.
GR-1~\citep{GR1} and GR-2~\citep{cheang2024gr2} use future image prediction as an auxiliary task to enhance the policy’s visual representations.
CLOVER~\citep{bu2024clover}, Seer~\citep{tian2025seer}, and $\mathcal{F}_1$~\citep{lv2025f1} leverage future image prediction to guide action generation in an inverse-dynamics-like manner.
As video generation models have advanced, it has become increasingly popular to incorporate a pretrained video generation model to guide action execution, as in VPP~\citep{hu2024vpp} and Genie Envisioner~\citep{liao2025genie_envisioner}.
Despite these advancements, these policies are often sensitive to video generation quality and lack the semantic reasoning capabilities inherent in MLLMs. In contrast, InternVLA-A1 integrates the MLLM with future visual state prediction, thereby 
providing semantically grounded guidance and enabling stronger adaptability in the presence of video prediction errors.

In \textbf{training data}, real-world robot data, simulated synthetic data, and human video data are widely used.
Real-world datasets~\citep{walke2023bridgedata, droid, bu2025agibotworld, wu2024robomind} capture realistic physical dynamics essential for end-to-end learning. RT-1~\citep{brohan2022rt1} and RT-2~\citep{zitkovich2023rt2} were trained on approximately 130k real-robot demonstrations.
And the Open X-Embodiment dataset~\citep{Neill2024oxe} aggregates one million episodes from multiple sources. 
$\pi_0$~\citep{black2024pi_0} further scales the amount of real-robot training data to 10,000 hours. 
However, further expanding real-robot datasets by another order of magnitude remains prohibitively expensive and operationally inefficient.
Simulation-based synthetic data~\citep{gu2023maniskill2,gao2025genmanip,robotwin2.0,huang2025roboground,james2020rlbench,mees2022calvin,hua2024gensim2,im,mandlekar2023mimicgen} can fully exploit rich scene and object assets, and use domain randomization to enrich sample diversity and cover long-tail scene variations. And importantly, It provides a low-cost, efficient route for data collection.
Robocasa~\citep{robocasa2024} collects data by teleoperating robots in simulation and then performing sample augmentation. Its initial data collection still requires manual effort. GraspVLA~\citep{GraspVLA} can automatically generate robot trajectories, but its skills are limited to grasping. More recently, InternData-A1~\citep{tian2025interndata_a1} proposes an automated robot trajectory synthesis pipeline that covers the manipulation of rigid, articulated, deformable, and fluid objects, and constructs a large-scale simulated demonstration dataset with over 630k trajectories and 7,433 hours of data spanning diverse embodiments, skills, tasks, and scenes. 
There is a vast amount of video data on the internet that remains underexplored for its potential to benefit the pretraining of manipulation policies.
Human videos~\citep{egodex,damen2020epic,grauman2024ego4d,goyal2017something} provide rich visual priors without costly teleoperation, proving effective for human-to-robot transfer~\citep{emergence_h2r, bi2025h_rdt}.
For instance, Ego4D~\citep{grauman2024ego4d} contains 3,670 hours of daily activities.
Tailored for manipulation, EgoDex~\citep{egodex} curates 829 hours of egocentric dexterous manipulation footage with paired 3D hand and finger tracking.
Some studies~\citep{cheang2024gr2, motus} have incorporated human video data to enrich the diversity of pre-training datasets.
In this work, we formulate a data recipe mixing real-world datasets, simulated data, and human videos, demonstrating that this combination enhances data diversity and effectively closes the sim-to-real gap.

\begin{figure}[t]
    \centering
    \includegraphics[width=\textwidth]{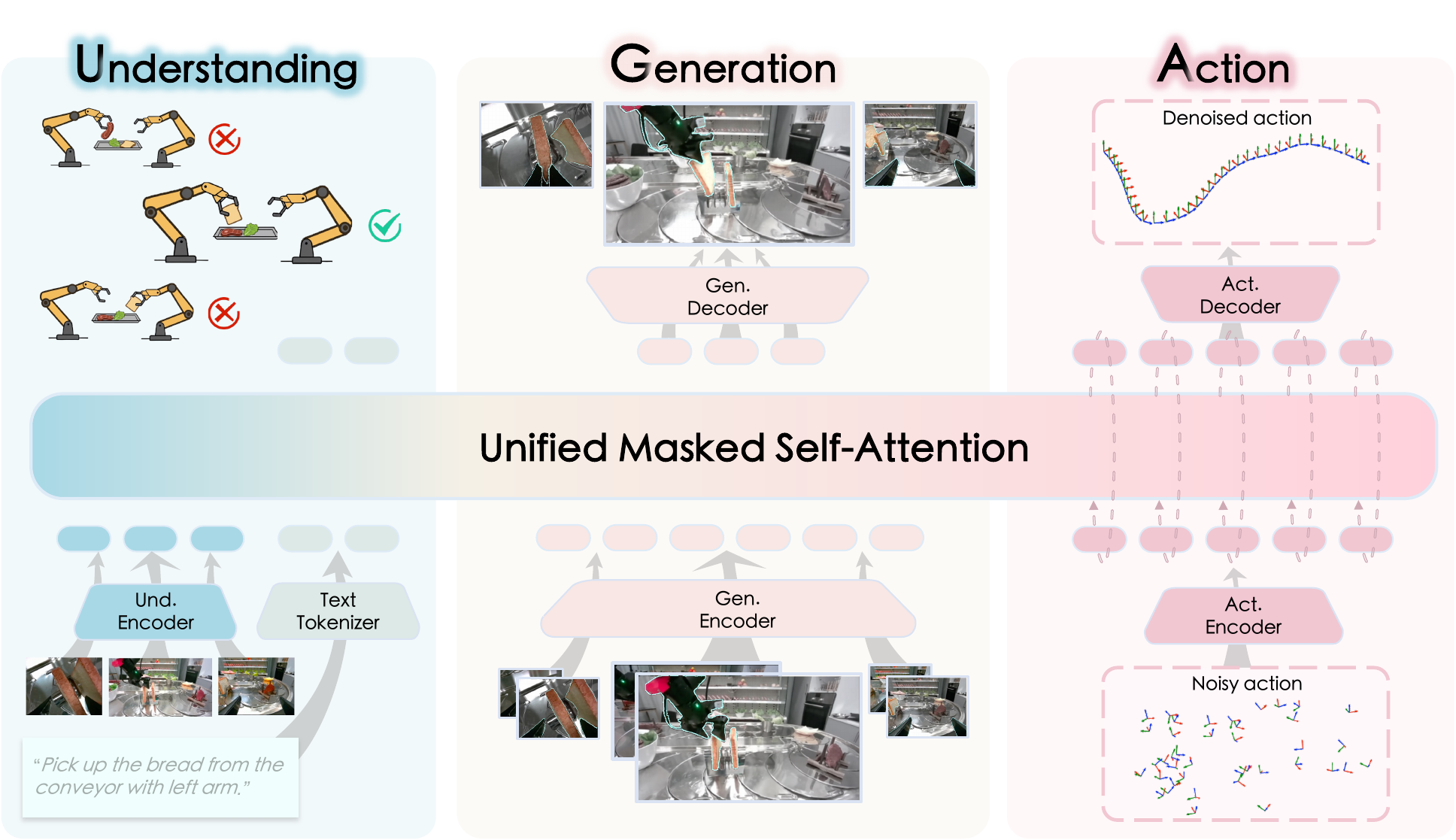} 
    \caption{
        \textbf{Framework of \modelname}. The architecture comprises three experts: (1) an \textbf{understanding expert} that encodes scene context from image and text inputs; (2) a \textbf{generation expert} that predicts future visual states and task dynamics; and (3) an \textbf{action expert} that integrates the encoded scene context with these predictive dynamics to synthesize control commands via Flow Matching. This tripartite design enables adaptive manipulation under scene variations.
    }
\label{figure:method_internvla}
\end{figure}

\section{InternVLA-A1: Unified Understanding-Generation-Action Framework}
\label{sec:method}

This section presents the design of \modelname. We first present an overview of the architecture, followed by a description of its components, optimization objectives, and implementation details.

\subsection{Architecture Overview}
The Mixture-of-Transformers (MoT) architecture, recently widely adopted in unified multimodal large language models~\citep{deng2025bagle}, demonstrates strong performance across both understanding and generation tasks. Drawing inspiration from these unified paradigms, \modelname adopts the MoT architecture to seamlessly integrate scene understanding, visual foresight, and action execution within a single framework.

As illustrated in \Cref{figure:method_internvla}, \modelname employs three experts in a unified pipeline. The understanding expert first processes multimodal inputs to capture the environmental context. These representations then inform the generation expert, which simulates the task's evolution by predicting future visual states. Finally, the action expert combines these predictive dynamics with the semantic context, utilizing Flow Matching to produce precise robot control commands.

\subsection{Core Components}
Inspired by the success of multimodal large language models, all three experts in \modelname adopt a decoder-only transformer architecture.

\noindent\textbf{Understanding Expert.} The understanding expert directly adopts the architecture of existing MLLMs. In this implementation, we employ InternVL3 or Qwen3-VL, distinguished by their native multimodal capabilities and strong alignment between language and vision. We adhere to the processing pipeline of the base MLLMs: the multi-view observation at time $t$, denoted $o_t$, is encoded into visual tokens via the integrated vision encoder, while language instructions $l$ are converted into text tokens using the text tokenizer. These visual and text tokens are subsequently processed by the transformer blocks of understanding expert, to form contextual embeddings $h_{\mathrm{und}} = f_{\mathrm{und}}(l,o_t)$. These embeddings serve as a shared context memory that is made accessible to downstream experts through masked self-attention, enabling the generation and action experts to attend to semantic scene context when predicting future latents and control commands.
% 
% The visual observation $o_t$ is encoded using the integrated vision encoder of the vision language model, following its established methodology.

\noindent\textbf{Generation Expert.} While video generation models have seen substantial progress, applying them to manipulation policies presents a significant challenge due to the requirements for high-frequency real-time inference. Mainstream image or video generation architectures, whether based on diffusion~\citep{rombach2022high,blattmann2023stable} or next-token prediction~\citep{sun2024Llamagen}, are typically too computationally intensive for end-to-end control. Even optimized solutions such as SANA-Sprint~\citep{chen2025sana} require 0.16 seconds per generation on one RTX 4090 GPU, restricting control frequencies to no more than 6Hz. Recent attempts~\citep{motus, dreamzero2025} leverage large-scale pretrained video foundation models and tailor them for action control; however, these methods lack sufficient real-time performance for high-dynamic scenarios. A notable example is DreamZero~\citep{dreamzero2025}: even after a 38$\times$ speedup from extensive engineering optimization, it attains only 7Hz on a GB200 GPU, presenting a significant barrier to robotic edge deployment and real-time inference.
Instead, we choose to implement a lightweight generation module. 
In our preceding work $\mathcal{F}_1$~\citep{lv2025f1}, we adopted a next-resolution paradigm for generating visual foresight. Although effective, this approach required iterative forward passes, which compromised the real-time inference capabilities essential for VLA tasks.
In this work, we have pursued a more efficient and effective design strategy.

\begin{figure}[t]
    \centering
    \includegraphics[width=\textwidth]{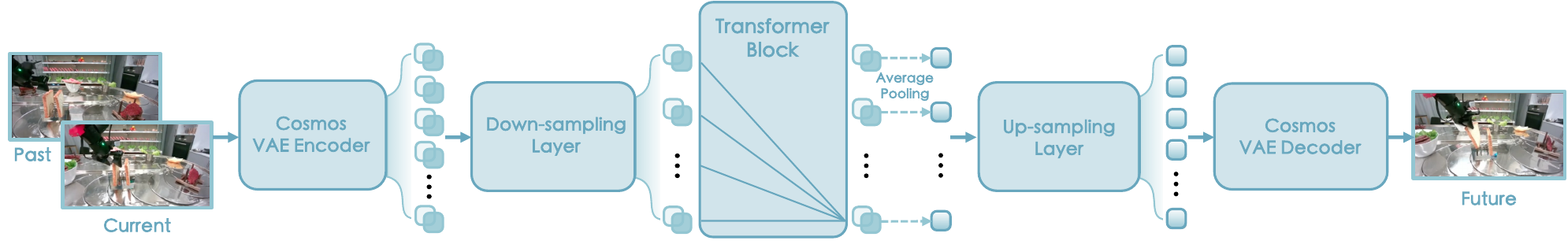} 
    \caption{
        \textbf{The internal architecture specifics of the Generation Expert}.
    }
\label{figure:specifics_gen_expert}
\end{figure}

% We process images from three views across two timestamps (the current frame and one historical frame) using the Cosmos CI$8\times8$ continuous VAE image tokenizer. With input images resized to $256 \times 256$, each is initially encoded into a $32 \times 32$ latent feature map. Directly feeding these latent tokens into the generation expert would result in an excessive sequence length, hindering both inference efficiency and training convergence.
% %
% To mitigate this, we apply a convolutional layer to compress the spatial dimensions of each latent feature to $4 \times 4$ (representing each image with just 16 tokens) and use a projector to align them with the transformer's hidden dimension. These compressed tokens are then input into the generation expert, where they attend to the prefix tokens $h_{\mathrm{und}}$ (cached as K/V at inference) via multiple layers of masked self-attention. Their hidden states are pooled over time to form the predicted future latent $\hat{z}_{t+m}$, supervised against the Cosmos encoding $z_{t+m}$ of frame $t+m$.
% %
% Finally, the predicted future latent are upsampled back to the original VAE latent dimensions using a deconvolutional layer and a projector. These restored features are fed into the Cosmos decoder to reconstruct the predicted images.

The detailed architectural design of the Generation Expert is illustrated in \Cref{figure:specifics_gen_expert}. Specifically, we expand upon three key specifics: Input Tokenization, Token Compression, and Parallel Decoding.

\textit{Input Tokenization}: the Generation Expert takes as input images from three perspectives, including a head view and two wrist views. To capture temporal dynamics, we sample frames from two timestamps: the current timestep $t$ and a historical timestep $t-15$. This results in a total of 6 input images, each resized to a resolution of $256 \times 256$. 
Inspired by the unified multimodal model Janus Pro~\citep{chen2025januspro}, we adopt a decoupled visual encoding strategy to address the divergent requirements of understanding and generation. Unlike understanding tasks demanding high-level semantic abstraction typically captured by ViT-based encoders, generation tasks require preserving fine-grained spatial structure and pixel-level fidelity. Accordingly, our generation expert employs a VAE-based tokenizer, widely employed in image and video generation for its ability to compress visual data into a latent space optimized for high-quality reconstruction. Specifically, we utilize the COSMOS CI8$\times$8 continuous VAE image tokenizer~\citep{agarwal2025cosmos} to encode these input images. In its raw form, each image is encoded into a $32 \times 32$ latent grid, resulting in 1,024 tokens per image. Directly feeding these $1,024 \times 6$ latent tokens into the generation expert would result in an excessive sequence length, hindering both inference efficiency and training convergence. Therefore, token compression is required.

\textit{Token Compression}: to address the sequence length challenge, we implement a token compression mechanism. We apply a convolution layer with an $8 \times 8$ kernel to downsample the latent grid, reducing the representation of each image to $4 \times 4$ (16 tokens). Consequently, the input sequence for the 6 images (3 views $\times$ 2 timestamps) is compacted to just 96 tokens. These tokens are then passed through a projector to align with the transformer blocks' hidden dimension.
Within the transformer blocks, tokens from the Generation Expert attend to both themselves and the Understanding Expert's tokens via a unified masked self-attention mechanism, utilizing the preceding Understanding Expert tokens as the KV cache. Subsequently, the processed tokens undergo parallel decoding to generate future images.

\textit{Parallel Decoding}: following the transformer processing, the output remains a sequence of 96 tokens. We apply temporal average pooling along the time axis to aggregate information from the two timestamps, resulting in 48 tokens representing the three views (16 tokens per view). These tokens are processed by a projector and subsequently upsampled via a deconvolution layer back to a $32 \times 32$ grid. Finally, the COSMOS VAE decoder reconstructs the predicted future frames (corresponding to $t+15$).
Notably, our approach avoids auto-regressive next-token prediction. Instead, we employ a single-forward parallel decoding strategy, generating all tokens for the future frames simultaneously. We find that this non-autoregressive approach is not only computationally efficient but also sufficient to provide effective visual guidance for the subsequent action execution.

\noindent\textbf{Action Expert.}
Conditioned on the latent features produced by the understanding and generation experts, together with the proprioceptive state $q_t$, the action expert predicts a target action chunk $\hat{a}_{t:t+k}$.
We adopt a flow matching objective to train the VLA model. 

\noindent\textbf{Attention Mechanism.} We implement a blockwise attention mask over the concatenated token streams of the understanding, generation, and action experts. A cumulative segment mask enforces a strict information flow understanding $\rightarrow$ generation $\rightarrow$ action: tokens in a later block can attend to all earlier blocks, while earlier blocks cannot attend forward. 
Within the transformer blocks of both the understanding and generation experts, the tokens are fully bidirectionally attended.
The tokens processed by the action expert's transformer blocks are split into a state token and action tokens. The state token attends only to itself and the tokens from earlier blocks, while action tokens attend to the state token, previous block tokens, and each other.

\subsection{Optimization Objectives}

Our training process consists of two sequential stages: \textbf{Pre-training} and \textbf{Post-training}. Although these two stages utilize distinct data sources and training hyperparameters (detailed in Table~\ref{tab:training_hyperparameters}), they share a unified optimization framework and identical objectives. Throughout both stages, we jointly optimize the model for two key objectives: visual foresight generation and action prediction.
Let $\mathcal{D}_1$ denote the training corpus for visual foresight (including human video data without action labels), and $\mathcal{D}_2$ denote the subset with action annotations. We define $\xi_1 = (o_{t-m}, o_t, o_{t+m}, l)$ and $\xi_2=(a_{t:t+k}, o_{t-m}, o_t, q_t, l)$ as the corresponding training tuples.

\noindent\textbf{(1) Visual Foresight Generation.} 
To endow the model with predictive capabilities about future visual states, we train the generation expert to forecast the latent representation of a future frame. Let $\phi_{\mathrm{cosmos}}$ denotes the COSMOS VAE encoder and $z_t=\phi_{\mathrm{cosmos}}(o_t)$ denotes the COSMOS latent feature. Conditioned on current and historical observations $\{o_{t-m}, o_t\}$, as well as the understanding prefix $h_{\mathrm{und}}$, the generation expert predicts the latent feature $\hat{z}_{t+m}$ for the future timestamp $t+m$. The prediction is supervised by the ground truth COSMOS latent feature $z_{t+m}$. We minimize the following objective:
\begin{equation}
\mathcal{L}_{\mathrm{gen}} = \mathbb{E}_{\xi_1} \left[ \left\| f_{\mathrm{gen}}(z_{t-m}, z_t; h_{\mathrm{und}}) - \mathrm{sg}[z_{t+m}] \right\|^2 \right]
\end{equation}
where $f_{\mathrm{gen}}$ denotes the generation expert, and $\mathrm{sg}[\cdot]$ indicates the stop-gradient operation.
This objective compels the model to internalize physical dynamics, creating a robust prior for action.

\noindent\textbf{(2) Flow Matching-based Action Prediction.}
We employ a flow matching framework for action learning.
This approach formulates the action generation process as learning continuous transformation pathways from noise to expert demonstrations, offering superior handling of multi-modal action distributions compared to direct regression.
Formally, given $\xi_2 \sim \mathcal{D}_2$, we sample time steps $\tau \sim \text{Beta}(1.5, 1.0)$ and construct interpolated action chunks $a_{t:t+k}^\tau = (1 - \tau)\epsilon + \tau a_{t:t+k}$, where $\epsilon \sim \mathcal{N}(0,I)$ represents Gaussian noise. 
The model learns a velocity field $v_\theta$ that transports noisy samples toward target actions:

\begin{equation}
  \mathcal{L}_{\mathrm{act}} = \mathbb{E}_{\xi_2} \Big[ \big\| v_\theta(q_t, a_{t:t+k}^\tau; h_{\mathrm{und}}, h_{\mathrm{gen}}) - (a_{t:t+k} - \epsilon) \big\|^2 \Big],
\end{equation}
where $q_t$ denotes the proprioception state at time $t$, and $h_{\mathrm{und}}$ and $h_{\mathrm{gen}}$ are the contextual conditioning features produced by the understanding expert and the generation expert, respectively.
During inference, sampling from the learned policy distribution is achieved by solving an ODE: starting from Gaussian noise $\epsilon \sim \mathcal{N}(0, I)$, we iteratively apply the Euler update:
\begin{equation}
   a_{t:t+k}^{\tau + \Delta\tau} = a_{t:t+k}^{\tau} + \Delta\tau \cdot v_\theta(q_t, a_{t:t+k}^\tau; h_{\mathrm{und}}, h_{\mathrm{gen}}),
\end{equation}
where $\tau$ progresses from $0$ to $1$ over $K$ steps with step size $\Delta\tau=1/K$.

\noindent\textbf{Loss Function.}
The total training objective is a weighted summation of two loss components:
\begin{equation}
  \mathcal{L}_{\mathrm{total}} = \lambda \cdot \mathcal{L}_{\mathrm{gen}} +  \mathcal{L}_{\mathrm{act}}
\end{equation}
where $\lambda$ is a hyperparameter balancing the two objectives. 
This joint optimization enforces representational consistency across modalities, enables implicit causal modeling of action-environment dynamics, and facilitates cross-modal knowledge transfer for enhanced generalization.

\looseness=-1

\subsection{Implementation Details}

\noindent\textbf{Model configurations and parameters.} We instantiate our model with two scales: InternVLA-A1 (2B) and InternVLA-A1 (3B). Both are built upon MLLM backbones and expanded into a unified system via the Mixture-of-Transformers (MoT) architecture.  Specifically, InternVLA-A1 (2B) utilizes InternVL3-1B as the understanding expert. Its generative and action experts are derived from the transformer blocks of Qwen2.5—the underlying LLM of InternVL3. 
InternVLA-A1 (3B) employs Qwen3-VL-2B as the foundation, with its generative and action experts derived from the Qwen3 transformer blocks. Detailed configurations and parameters are provided in Table~\ref{tab:model_parameters}.

Regarding inference efficiency, both InternVLA-A1 (2B) and InternVLA-A1 (3B) run at approximately 13 Hz with \texttt{torch.compile} on one RTX 4090 GPU. 
Notably, InternVLA-A1 (2B) does not exhibit lower latency than InternVLA-A1 (3B) despite having fewer total parameters. This is because the InternVL3 backbone in InternVLA-A1 (2B) requires a higher input resolution $448 \times 448$ compared to the $224 \times 224$ input used by the Qwen3-VL backbone in InternVLA-A1 (3B). Consequently, the computational cost of processing longer visual token sequences in InternVLA-A1 (2B) offsets its parameter advantage, leading to comparable overall speeds.

\begin{table}[h]
\centering
\caption{Model configurations and parameters for InternVLA-A1. The inference speed in frame per second (FPS) is evaluated on NVIDIA RTX 4090 GPU.}
\small
\begin{tabular}{l c c c c c}
\toprule
\textbf{Model Variant} & \textbf{\#Param.} & \textbf{Und. Expert} & \textbf{Gen. Expert} & \textbf{Act. Expert} & \textbf{FPS} \\
\midrule
InternVLA-A1 (2B) & 1.8B & InternVL3 (0.94B) & Qwen2.5 (0.36B) & Qwen2.5 (0.36B) & $\sim$13 Hz\\
InternVLA-A1 (3B) & 3.2B & Qwen3-VL (2.13B) & Qwen3 (0.44B) & Qwen3 (0.44B) & $\sim$13 Hz\\
\bottomrule
\end{tabular}
\label{tab:model_parameters}
\end{table}

\noindent\textbf{Training Protocol and Hyperparameters.}
We train InternVLA-A1 in two stages: a large-scale pre-training stage on heterogeneous datasets, followed by a task-specific post-training stage.
During pre-training, we optimize the model using AdamW with a constant learning rate schedule for 700K steps. For post-training, we adopt a lower learning rate with a warmup and decay schedule to stabilize adaptation to downstream tasks. 
Detailed hyperparameter settings for pre-training and post-training are provided \Cref{tab:training_hyperparameters}. In addition, 
the hyperparameter $\lambda$, which balances the two loss components, is set to 0.01.
The interval $m$ between the historical frame and the current frame, as well as between the future frame and the current frame, is set to 15.

\begin{table}[h]
\centering
\caption{Training hyperparameters for InternVLA-A1.}
\small
\setlength{\tabcolsep}{8mm}
\begin{tabular}{l c c}
\toprule
\textbf{Configuration} & \textbf{Pre-training} & \textbf{Post-training} \\
\midrule
Optimizer & AdamW & AdamW \\
Batch size & 512 & 128 \\
Learning rate & $5 \times 10^{-5}$ (constant) & $5 \times 10^{-5} \rightarrow 5 \times 10^{-6}$ \\
Warmup steps & -- & 2,000 \\
Decay steps & -- & 60,000 \\
Training steps & 700K & 60,000 \\
Optimizer betas & $\beta_1=0.9,\ \beta_2=0.95$ & $\beta_1=0.9,\ \beta_2=0.95$ \\
Optimizer epsilon & $1 \times 10^{-8}$ & $1 \times 10^{-8}$ \\
Weight decay & 0.01 & 0.01 \\
Gradient clipping & 1.0 & 1.0 \\
Model precision & bfloat16 & bfloat16 \\
\bottomrule
\end{tabular}
\label{tab:training_hyperparameters}
\end{table}

\noindent\textbf{Load-balanced Parallel Training.}
InternVLA-A1 is trained on a mixture of heterogeneous datasets over 692 frames. At this scale, naively instantiating the complete
datasets on every worker like LeRobot codebase~\citep{cadene2024lerobot} can trigger out-of-memory issues and exacerbate I/O contention. We therefore adopt
Load-balanced Parallel Training (LPT), a distributed data-loading strategy that assigns datasets to workers
to achieve both scalability and statistically well-behaved sampling.

Let $\{D_i\}_{i=1}^n$ denote the set of training datasets,and let $s_i$ be a lightweight proxy of the size of $D_i$
(e.g., the number of frames). LPT computes an assignment $\pi:\{1, \dots, n\}\rightarrow\{1, \dots, K\}$ that maps each
dataset to one of $K$ workers, subject to two desiderata: coverage (each worker receives at least one dataset)
and balance (the total assigned size per worker is approximately uniform). In practice, we employ a simple
greedy load-balancing rule that iteratively assigns the next dataset to the currently least-loaded worker:
\[
\pi(i) \;=\; \text{argmin}_{k\in\{1, \dots, K\}} \sum_{j:\, \pi(j)=k} s_j \,,
\]
with datasets processed in descending order of $s_i$. This procedure is efficient, deterministic, and empirically
yields near-uniform per-worker loads.

LPT improves robustness in large-scale training by (i) reducing per-worker memory pressure since each worker materializes only a subset of datasets, and (ii) mitigating implicit re-weighting effects that would otherwise arise when workers traverse datasets with highly heterogeneous sizes at different rates. 
When the number of datasets is smaller than the number of workers, we allow controlled replication to avoid idle workers. Replicated datasets are assigned to different workers with independent random seeds and load-aware placement, such that no worker is disproportionately dominated by a small dataset. While replicas may sample from overlapping episode pools, this strategy empirically approximates uniform effective sampling by equalizing per-worker data throughput, thereby mitigating the implicit re-weighting effects introduced by dataset size heterogeneity.

% \noindent\textbf{Training and Inference.}

\section{Data Corpus}
\label{sec:data_corpus}

\subsection{Pre-training data recipe}
We pretrain the model on a mixture of heterogeneous data sources including synthetic simulation data, real-world robot data, and human videos. The pre-training data recipe is shown in Table~\ref{tab:pretrain_sampling_weights}. During pretraining, we interleave trajectories from different sources using configurable sampling weights.

\begin{table}[h]
    \centering
    \caption{Data mixture used for pretraining.}
    \small
    \begin{tabular}{l|c|c|c}
        \toprule
        \textbf{Data source} & \textbf{Type} & \textbf{Num.\ frame} & \textbf{Sampling weight} \\
        \midrule
        InternData-A1        & Sim.   & 396M & 0.64 \\
        RoboTwin            & Sim.   & 17M  & 0.08 \\
        AgiBot-World (Beta)  & Real   & 206M & 0.18 \\
        RoboMind            & Real   & 5M   & 0.02 \\
        EgoDex              & Human  & 68M  & 0.08 \\
        \bottomrule
    \end{tabular}
    \label{tab:pretrain_sampling_weights}
\end{table}

\subsection{Simulated synthetic data}

We incorporate  our prior work, InternData-A1~\citep{tian2025interndata_a1}, a large-scale synthetic robot dataset that is among the most diverse and comprehensive to date.
The dataset contains over 630k trajectories and 7,433 hours of data spanning 4 embodiments, 18 skills, 70 tasks, and 227 scenes, covering manipulation of rigid, articulated, deformable, and fluid objects.
It is generated via a highly autonomous, fully decoupled, and compositional simulation pipeline, enabling long-horizon skill composition, flexible task assembly, and support for heterogeneous embodiments with minimal manual tuning.
InternData-A1 is the first to demonstrate that synthetic-only data can match the performance of large-scale real-world datasets when pre-training VLA models, achieving comparable results to the strongest closed-source real-world $\pi$-dataset~\citep{black2024pi_0}.
Moreover, models trained on InternData-A1 exhibit strong zero-shot sim-to-real transfer on several challenging tasks.
With InternData-A1's synthesis pipeline and system optimization enabled by the Nimbus~\citep{he2026nimbus}, it generates 209.7 hours of simulation data per day on 8 RTX 4090 GPUs.
As the foundation of our data corpus, InternData-A1 provides rich diversity in trajectories, objects, and environments.
We select InternData-A1 as the foundation of our pre-training corpus due to its exceptional sample diversity and proven efficacy in pre-training VLA models. Additionally, we incorporate the simulation dataset proposed in RoboTwin~\citep{robotwin2.0}.

\subsection{Real-world robot data}
While synthetic data excels in scalability, real-world demonstrations remain essential for capturing nuanced physical dynamics and bridging the sim-to-real gap.
When selecting real-world demonstration data for our pre-training corpus, we prioritize datasets that exhibit large-scale trajectory coverage, diverse task distributions, and high-quality teleoperation.
Based on these criteria, we incorporate the open-source AgiBot-World dataset~\citep{bu2025agibotworld} and RoboMind~\citep{wu2024robomind}.
By incorporating these large-scale real-world demonstration datasets into our pre-training corpus, we benefit from its high-quality and diverse demonstrations, which complement our synthetic data and help bridge the sim-to-real gap.

\subsection{Egocentric human video}
We additionally incorporate human video data into our data corpus.
To align with robot manipulation scenarios, we prioritize egocentric videos that share a similar viewpoint and feature diverse interactions in robot-like environments.
Specifically, we utilize the EgoDex~\citep{egodex} dataset, a large-scale collection focused on egocentric dexterous manipulation.
It comprises 829 hours of footage spanning over 200 tasks. Notably, we exclude human action labels during pre-training.
Rich in human-hand-object interactions, this data is instrumental for the generation expert, enabling it to capture the nuanced dynamics and diverse manipulation skills inherent in the real world. The human video data can also be further expanded to incorporate large-scale datasets like Ego4D~\citep{grauman2024ego4d} and EPIC-KITCHENS~\citep{damen2020epic}.

\section{Experiments}
\label{sec:exp}

To evaluate the effectiveness of our proposed \modelname, we conduct extensive experiments on 12 real-world tasks and a simulation benchmark. We first compare \modelname with existing leading VLA models.
Following this, we conduct a series of ablation studies.

\subsection{Evaluation Configuration}
\paragraph{Hardware \& platform.}
We benchmark the policies on three physical robot embodiments: \textbf{Agibot Genie-1}, \textbf{ARX Lift-2}, and \textbf{ARX AC One}. These platforms cover diverse bimanual manipulation capabilities and execution characteristics, enabling us to assess real-robot performance across both long-horizon and contact-rich behaviors under consistent sensing and control setups. All evaluations are performed with the same deployment pipeline across the three robots.

\noindent\textbf{Task setup.}
We design a real-world task suite comprising \textbf{ten} static manipulation tasks and \textbf{two} dynamic task families:
\begin{itemize}
    \item \textbf{Static manipulation tasks:} \textit{Zip Bag}, \textit{Unscrew Cap}, \textit{Make Sandwich}, \textit{Operate Oven}, \textit{Sort Parts}, \textit{Sort Rubbish}, \textit{Sweep Trash}, \textit{Wipe Stain}, \textit{Place Markpen}, \textit{Place Flower}.
    \item \textbf{Dynamic manipulation tasks:} \textit{Express Sorting tasks} and \textit{In-motion Ingredient Picking tasks}.
\end{itemize}
Overall, the tasks span articulated-object interaction and contact-rich manipulation (e.g., Operate Oven and Unscrew Cap), long-horizon bimanual manipulation (e.g., Zip Bag and Make Sandwich), and, importantly, high-dynamics settings where the scene evolves and the target is moving during execution. The latter requires the policy to reason about near-future scene changes, so that actions can be generated with upcoming dynamics in mind rather than solely based on the current observation.

\noindent\textbf{Evaluation protocol.}
We report the average success rate over 30 rollouts per task. Specifically, each task is evaluated under 30 predefined settings (e.g., object placements and scene initializations within bounded ranges), with one trial per setting. Results are averaged across all rollouts to summarize performance across diverse conditions.

\subsection{Evaluation on the Static Manipulation Tasks}

\begin{figure}[t]
    \centering
    \includegraphics[width=\textwidth]{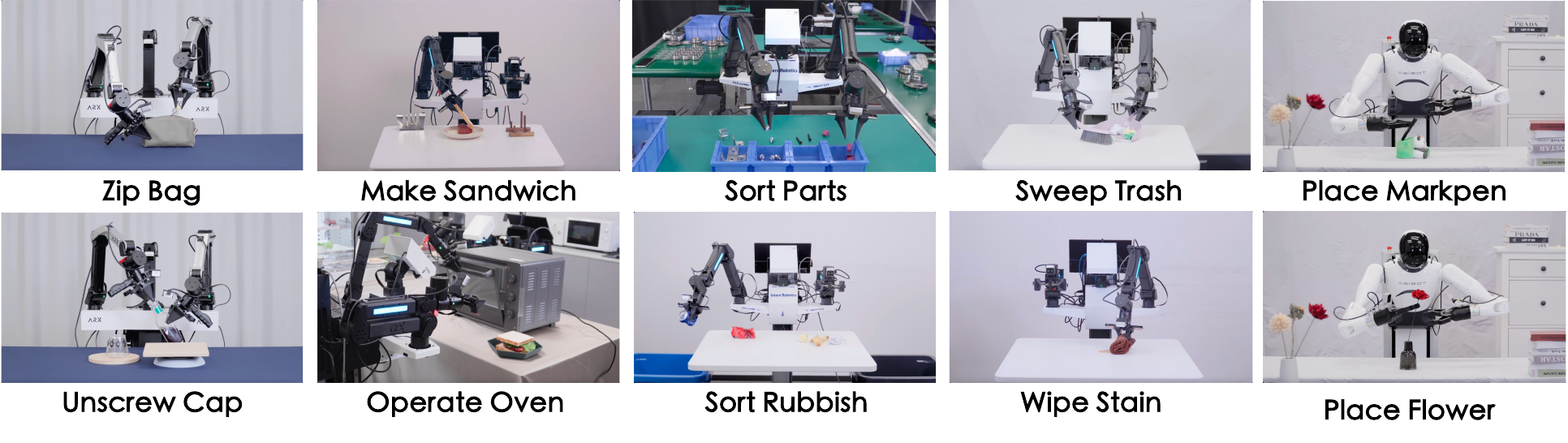} 
    \caption{
        \textbf{The experimental setting of real-world static tasks}. 
    }
\label{figure:general-purpose-tasks}
\end{figure}

To evaluate the general manipulation capabilities of \modelname against leading VLA models, we conducted a comprehensive evaluation across 10 diverse real-world static tasks (\Cref{figure:general-purpose-tasks}).

As shown in Table~\ref{table:results_general_purpose_tasks}, InternVLA-A1 clearly outperforms prior state-of-the-art models on real-world static manipulation tasks. Notably, our smaller InternVLA-A1 (2B) achieves an average success rate of 64.7\%, surpassing the 60.6\% average of the larger $\pi_0$ (3.3B) baseline. This result highlights the effectiveness of our architecture and joint training of simulated trajectories, real-robot demonstrations, and egocentric human videos, enabling a lighter model to outperform a larger counterpart.
When scaling to a comparable size, InternVLA-A1 (3B) delivers a clear performance gain, reaching an average success rate of 75.1\%, which is +14.5\% over $\pi_0$ (60.6\%) and +4.4\% over $\pi_{0.5}$ (70.7\%). It shows strong advantages on long-horizon tasks such as Make Sandwich (93.3\% vs. 73.3\% for $\pi_{0.5}$) and Operate Oven (86.7\% vs. 80.0\%), and remains top-tier on Sort Rubbish (97.3\%), matching the best baseline. On challenging manipulation tasks, InternVLA-A1 (3B) matches the strongest baseline on Unscrew Cap (66.7\%) and Sort Parts (53.3\%), while achieving substantial gains on Zip Bag (73.3\% vs. 60.0\% for $\pi_{0.5}$, and 40.0\% for $\pi_0$). These results confirm that InternVLA-A1 not only excels in high-level task planning but also possesses superior fine-grained control capabilities.

\begin{table*}[h!]
  \footnotesize
  \centering
  \caption{\textbf{Experimental results on real-world general-purpose tasks.} The best results are \textbf{bolded}. 
  } \label{table:results_general_purpose_tasks}

  \setlength{\tabcolsep}{4mm} %
  \begin{tabular}{cccccc}
    \toprule
    % ==================== 
    Method &
    Zip Bag &
    Unscrew Cap &
    Sort Parts &
    Make Sandwich &
    Sweep Trash
    \\
    \midrule
    GR00T N1.5  & 33.3  & 0.0 & 6.7  & 46.7 & 16.7 \\
    $\pi_0$  & 40.0 & \textbf{66.7} &  \textbf{53.3} & 66.7 & 43.3\\
    $\pi_{0.5}$  & 60.0 & \textbf{66.7} & \textbf{53.3} & 73.3 & 50.0 \\
    InternVLA-A1(2B)  & 66.7 & 33.3 &  46.7 & 73.3 & 63.3 \\
    InternVLA-A1(3B)  & \textbf{73.3} & \textbf{66.7} &  \textbf{53.3} & \textbf{93.3} & \textbf{66.7}\\
    \midrule
    \midrule
    % ====================
    Operate Oven &
    Sort Rubbish &
    Wipe Stain &
    Place Markpen &
    Place Flower &
    Average \\
    \midrule
     46.7 & 66.7 & 40.0 & 40.0 & 33.3 & 33.0\\
     73.3 & 96.0 & 73.3 & 53.3 & 40.0 & 60.6\\
     80.0 & \textbf{97.3} & \textbf{93.3} & \textbf{66.7} & \textbf{66.7} & 70.7 \\
     53.3 & \textbf{97.3} & 80.0 & \textbf{66.7} & \textbf{66.7} &  64.7\\
     \textbf{86.7} & \textbf{97.3} & 86.7 & \textbf{66.7} & 60.0 & \textbf{75.1}\\
    \bottomrule
  \end{tabular}
\end{table*}

\begin{figure}[t]
    \centering
    \includegraphics[width=\textwidth]{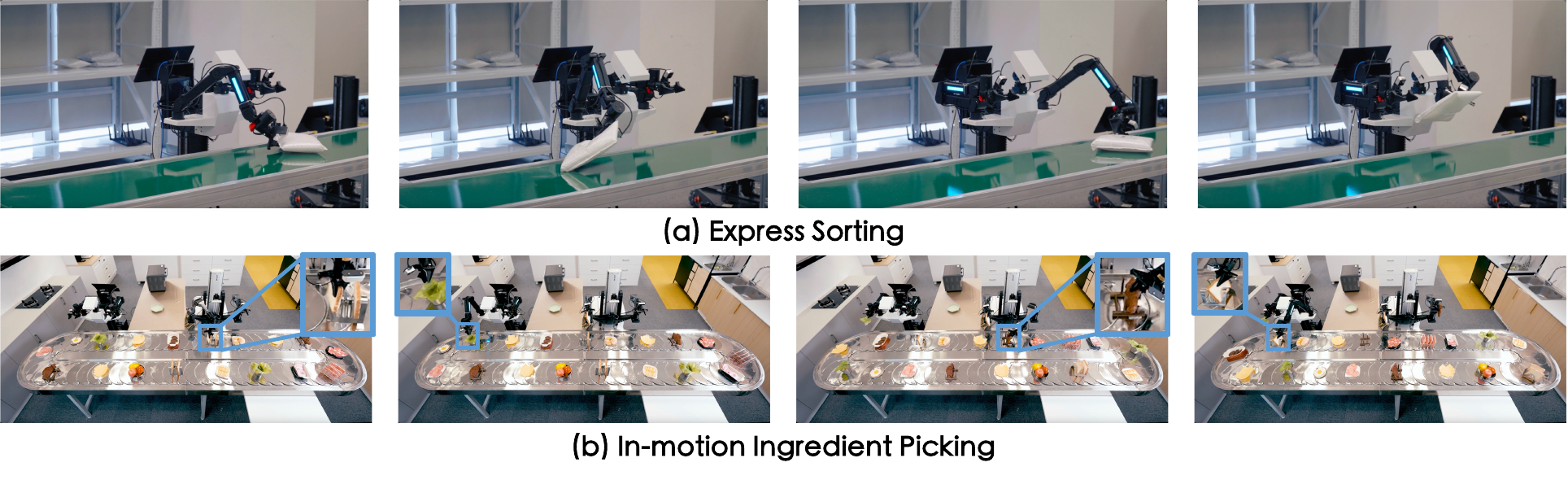} 
    \caption{
        \textbf{The experimental setting of real-world dynamic tasks:  (a) Express Sorting task, (b) In-motion Ingredient Picking task.}
    }
\label{figure:dynamic_tasks}
\end{figure}

\subsection{Evaluation on Dynamic Manipulation Tasks}

To assess robustness under environmental dynamics, we evaluate the policies on two challenging real-world dynamic manipulation tasks: \textit{Express Sorting} and \textit{In-motion Ingredient Picking}, where targets are moving during execution.
The experimental settings for the Express Sorting and In-motion Ingredient Picking tasks are depicted in \Cref{figure:dynamic_tasks}. Both involve long-horizon operations within dynamic environments.
In the Express Sorting Task, the robot must first determine if the package label is facing upwards. If the label is facing downwards, a four-step sequence is initiated: the right arm first executes a ``chasing'' grasp along the direction of the conveyor movement, followed by flipping the package, after which the left arm performs a ``waiting'' grasp. Finally, the package is lifted to present the label to the head-mounted camera. Conversely, if the label is facing upwards, the robot skips the flipping sequence and directly proceeds to the final two steps.
In the In-motion Ingredient Picking tasks, two robots coordinate to grasp the ingredients required to assemble a beef sandwich, consisting of two slices of bread, a steak, and a piece of lettuce.

\begin{figure}[h]
    \centering
    \includegraphics[width=0.8\textwidth]{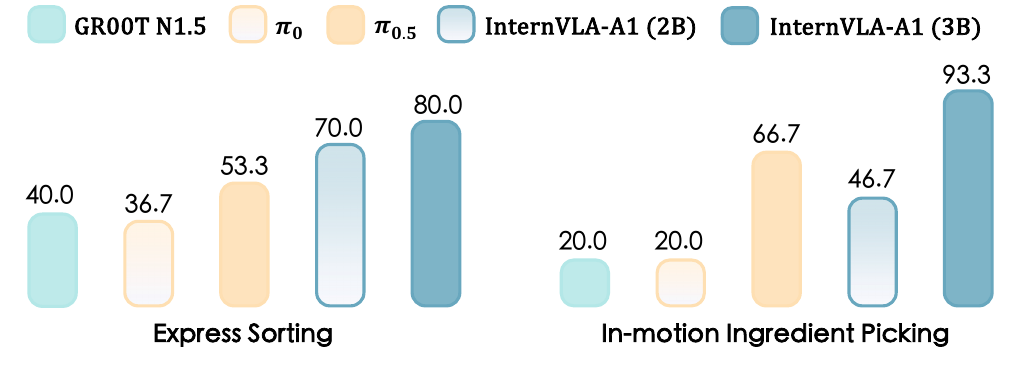} 
    \caption{
        \textbf{The experimental results of Express Sorting and In-motion Ingredient Picking tasks}. 
    }
\label{figure:results_dynamic_tasks}
\end{figure}

As shown in \Cref{figure:results_dynamic_tasks}, InternVLA-A1 establishes a clear lead over all baselines, surpassing even the  strongest prior VLA model $\pi_{0.5}$. Notably, the InternVLA-A1 (3B) variant demonstrates the most pronounced advantage. While InternVLA-A1 (2B) already performs strongly on Express Sorting (70.0\%), surpassing $\pi_{0.5}$ (53.3\%) by +16.7\% and substantially outperforming GR00T N1.5 (40.0\%) and $\pi_{0}$ (36.7\%), scaling to 3B pushes performance to 80.0\% on Express Sorting and 93.3\% on Ingredient Picking. Notably, on Ingredient Picking, a highly dynamic and timing-sensitive setting, InternVLA-A1 (3B) delivers a substantial gain over $\pi_{0.5}$ (93.3\% vs.\ 66.7\%, +26.6\%), while $\pi_{0}$ and GR00T N1.5 remain at 20.0\%. Similarly, on Express Sorting it improves over $\pi_{0.5}$ by +26.7\% (80.0\% vs.\ 53.3\%). Overall, InternVLA-A1 (3B) achieves an average success rate of 86.7\% across these two dynamic tasks. These results highlight that InternVLA-A1 is markedly more robust to environmental dynamics and motion-induced distribution shifts, translating foresight-guided decision making into reliable closed-loop control in dynamic scenarios.

\subsection{Evaluation on Simulation Benchmark}

% \begin{table*}[t!]
%   \footnotesize
%   \centering
%   \caption{\textbf{Evaluation on RoboTwin 2.0 Simulation Benchmark.} The best results are \textbf{bolded}. 
%   } \label{table:robotwin_sim_benchmark}
  
%   % 定义列格式：1个左对齐(Method) + 4组数据(每组2列) = 9列
%   % 这样上半部分填满，下半部分最后留空
%   \setlength{\tabcolsep}{3.5mm} % 调整列间距以适应页面
%   \begin{tabular}{ll|cc}
%     \toprule
%     % ==================== 上半部分 (前4个任务) ====================
%     \multicolumn{2}{c|}{\multirow{2}{*}{Method}} &
%     \multicolumn{2}{c}{Avg. (50 Tasks)} \\
%     \cmidrule(lr){3-4} % 横线覆盖第2到第9列
%      & & Easy & Hard\\
%     \midrule
%     \multicolumn{2}{c|}{$\pi_0$}  & 79.98\% & 79.50\% \\
%     \multicolumn{2}{c|}{$\pi_{0.5}$} & 86.76\% & 86.96\% \\
%     \multicolumn{2}{c|}{InternVLA-A1(3B)} & \textbf{89.40\%} & \textbf{89.64\%} \\
%     \bottomrule
%   \end{tabular}
% \end{table*}

\begin{figure}[h]
    \centering
    \includegraphics[width=0.8\textwidth]{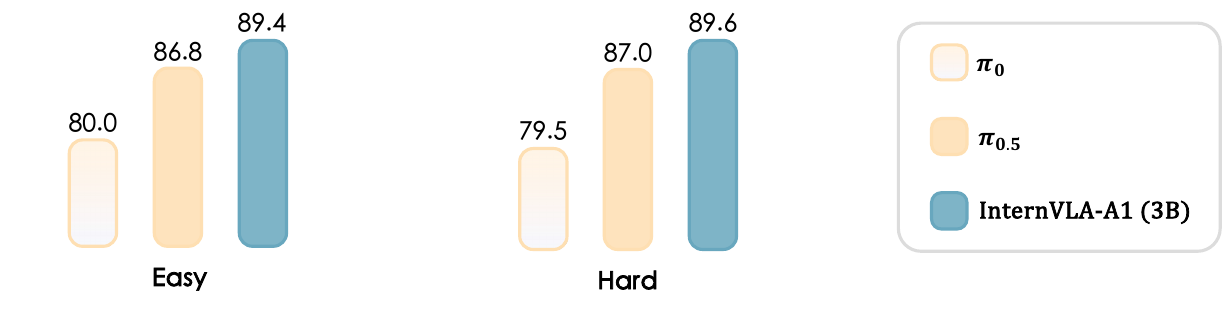} 
    \caption{
        \textbf{Evaluation on RoboTwin 2.0 Simulation Benchmark}. 
    }
\label{figure:results_sim_benchmark}
\end{figure}

We also evaluate \modelname on the RoboTwin 2.0~\citep{robotwin2.0} benchmark, covering 50 bimanual tasks under both Easy (clean) and Hard (domain randomized) settings.
All models are fine-tuned on a total of 27,500 demonstrations, consisting of 2,500 clean and 25,000 randomized episodes (i.e., 50 and 500 per task), with results averaged over 100 evaluation trials.
As shown in \Cref{figure:results_sim_benchmark}, InternVLA-A1(3B) significantly outperforms $\pi_0$ by margins of 9.4\% and 10.1\%, and surpasses the strong $\pi_{0.5}$ baseline by 2.6\% in both the Easy and Hard settings.

\subsection{Ablation Studies}

\begin{figure}[h]
    \centering
    \includegraphics[width=\textwidth]{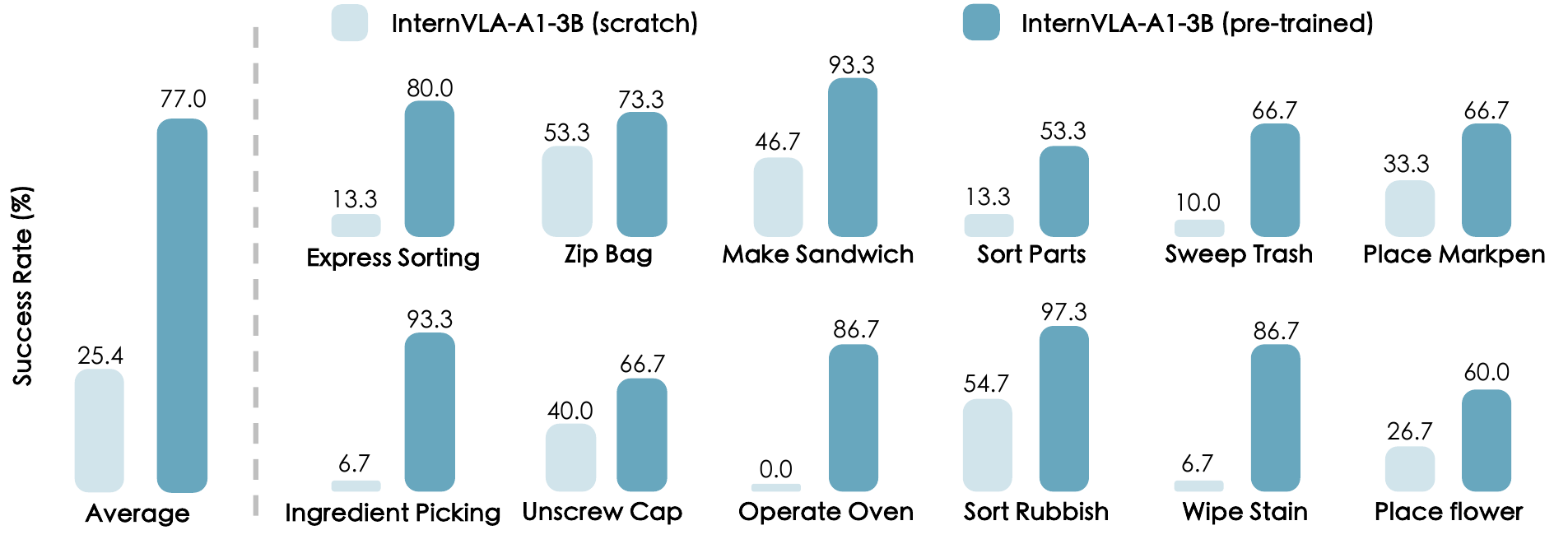} 
    \caption{
        \textbf{The ablation studies on pre-training}. 
    }
\label{figure:ablation_study_on_pretraining}
\end{figure}

\noindent\textbf{Impact of Pre-training.} As shown in Figure~\ref{figure:ablation_study_on_pretraining},  we assess the impact of pre-training by comparing our model with a version trained from scratch. The removal of the pre-training stage resulted in an overall performance drop of 51.6\%, with the average success rate falling from 77.0\% to 25.4\%. In severe cases, the baseline failed completely, while the pre-trained model retained high proficiency. This finding suggests that pre-training acts as a crucial inductive prior.

\noindent\textbf{Impact of Pre-training Datasets}. As shown in \Cref{table:ablation_study_on_dataset}, we evaluate the efficacy of different pre-training data sources.
Training on simulation data alone already yields strong results on RoboTwin 2.0 (88.3\%/88.5\% on Easy/Hard), but generalizes less well to real-world tasks (53.3\% on Place flower and 33.3\% on Sort parts). Incorporating human videos improves simulation performance (up to 89.4\%/89.3\%) and slightly benefits real-world generalization (specifically improving Sort parts to 40.0\%). Most notably, jointly pre-training on heterogeneous data sources (human videos, synthetic data, and real-world demonstrations) achieves the best overall performance, and substantially enhances real-world manipulation performance (reaching 60.0\% on Place flower and 53.3\% on Sort parts). This demonstrates the effectiveness of our joint training strategy.

\begin{table}[h!]
  \footnotesize
  \centering
  \caption{\textbf{Ablation studies on the pre-training dataset.}}
  \label{table:ablation_study_on_dataset}
  \setlength{\tabcolsep}{3.0mm}
  \begin{tabular}{l|cc|cc}
    \toprule
    Pre-training dataset &
    \multicolumn{2}{c|}{RoboTwin 2.0} &
    \multicolumn{2}{c}{Real-world tasks} \\
    \cmidrule(lr){2-3}
    \cmidrule(lr){4-5}
     & Easy & Hard & Place flower & Sort parts \\
    \midrule
    Sim. only
      & 88.3  & 88.5   & 53.3 & 33.3 \\
    Sim. + Human
      & \textbf{89.4} & 89.3 & 53.3 & 40.0 \\
    Sim. + Real + Human
      & \textbf{89.4} & \textbf{89.6} & \textbf{60.0} & \textbf{53.3} \\
    \bottomrule
  \end{tabular}
\end{table}

\begin{figure}[t]
    \centering
    \includegraphics[width=\textwidth]{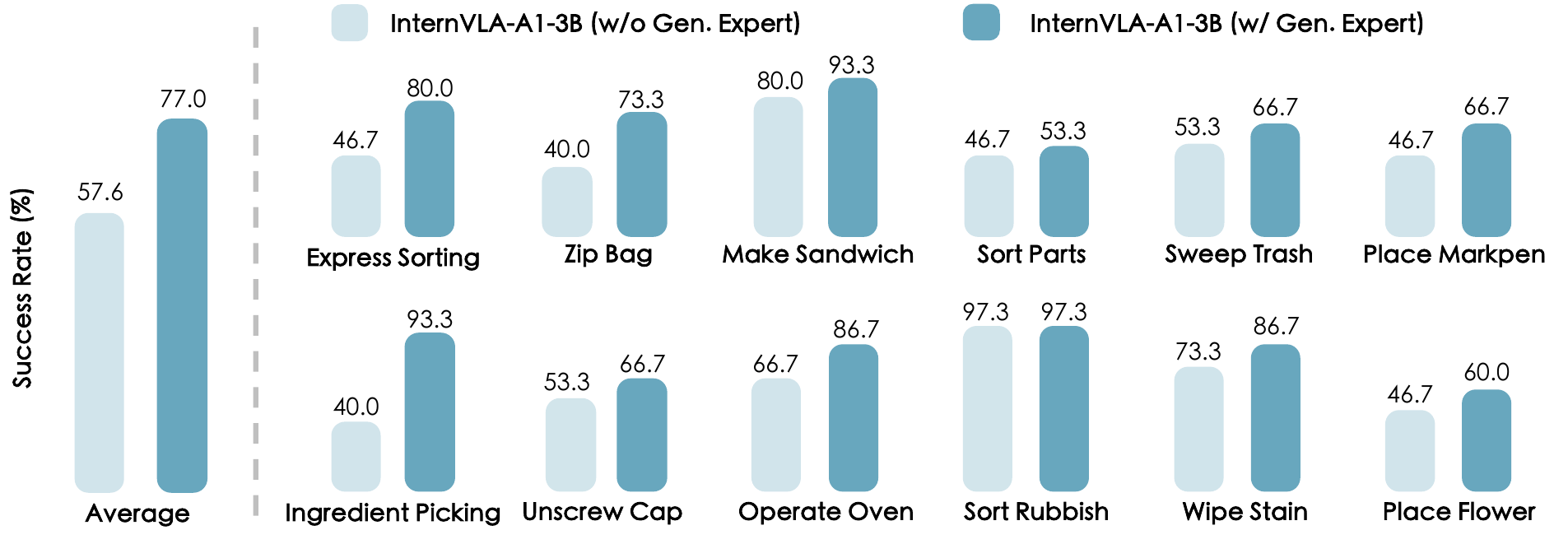} 
    \caption{
        \textbf{The ablation studies on generation expert}. 
    }
\label{figure:ablation_study_on_gen_expert}
\end{figure}

\noindent\textbf{Impact of Generation Expert.} To evaluate the contribution of the generation expert, we conduct a key ablation study by comparing InternVLA-A1 (3B) with the variant without the generation expert. The experimental results are presented in \Cref{figure:ablation_study_on_gen_expert}.
The results demonstrate that InternVLA-A1 (3B) outperforms the ablated version (without the generation expert) in 11 out of 12 real-world tasks, achieving performance gains ranging from 6.7\% to 53.3\%. Notably, removing the generation expert significantly reduces the average success rate from 77.0\% to 57.6\%, with the most substantial decline observed in dynamic manipulation tasks ( Express Sorting and In-motion Ingredient Picking). This ablation study validates the superiority of the proposed generation expert and the unified architecture integrating understanding, generation, and action.

\subsection{Visualization}

\begin{figure}[h]
    \centering
    \includegraphics[width=\textwidth]{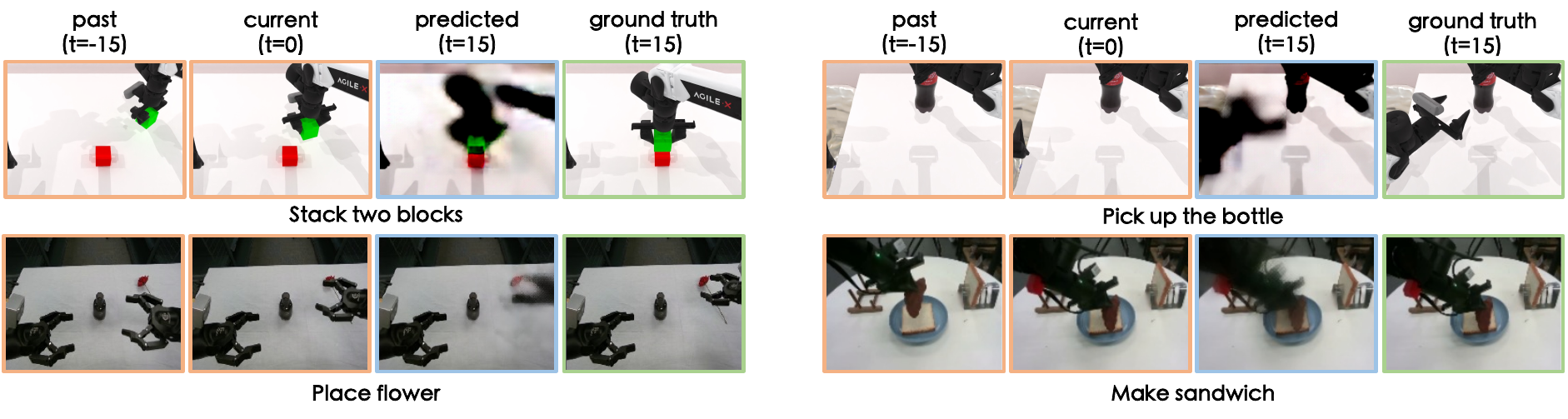} 
    \caption{
        \textbf{Visualization of future image prediction}. 
    }
\label{figure:visualization}
\end{figure}

The head-view future predictions generated by InternVLA-A1 are illustrated in \Cref{figure:visualization}. As shown in the samples, the predictions sacrifice some high-frequency visual details. This is a deliberate design choice to prioritize inference efficiency over pixel-level granularity. Despite this, the model accurately captures the essential motion trends and dynamics. We argue that when integrating visual foresight generation module into manipulation policies, high-frequency detail or visual clarity is secondary; what matters most is that the latent features encapsulate sufficient instructive information to guide action execution.

\section{Conclusion and Limitations}
\label{sec:conclusion}

In this work, we presented \modelname, a unified framework that integrates scene understanding, visual-foresight generation, and action execution through a Mixture-of-Transformers (MoT) design. This architecture couples semantic reasoning with dynamics prediction, and enables effective joint training on heterogeneous data sources (human videos, synthetic data, and real-world demonstrations). As a result, \modelname achieves consistent robustness across static manipulation, dynamic manipulation, and simulation benchmarks, with particularly strong gains in highly dynamic scenarios.

\noindent\textbf{Limitations.} Despite these advancements, two primary limitations remain. First, the understanding expert is not jointly trained with large-scale multimodal VQA data, which weakens general semantic reasoning and complex instruction following. Second, to ensure efficient inference for the visual foresight generation module, we compromised the fidelity of image prediction, limiting the granularity of generated future frames. We will address the limitations in future work.

\newpage

\bibliography{refs}

\appendix
\newpage

\section{Contributors}
\label{sec:contributors}

All contributors are listed in alphabetical order by their last names.

% \sloppy
% \subsection*{\textbf{\color{yellow}Core Contributors}}
% \mbox{Junhao~Cai\textsuperscript{1}}, \mbox{Yang~Li\textsuperscript{1}}, 
% \mbox{Haoxiang~Ma\textsuperscript{1}}, \mbox{Jiangmiao~Pang\textsuperscript{1\dag}}, \mbox{Zherui~Qiu\textsuperscript{1}}, \mbox{Yang~Tian\textsuperscript{1}}, \mbox{Jia~Zeng\textsuperscript{1\dag},
% \mbox{Hongrui~Zhu\textsuperscript{1}}}

{
\raggedright
\spaceskip=0.3em
\subsection*{\textbf{\color{yellow}Core Contributors}}
\mbox{Junhao Cai\textsuperscript{1}}, \mbox{Yang Li\textsuperscript{1}}, 
\mbox{Haoxiang Ma\textsuperscript{1}}, \mbox{Jiangmiao Pang\textsuperscript{1$\natural$}}, 
\mbox{Zherui Qiu\textsuperscript{1}}, \mbox{Yang Tian\textsuperscript{1}}, 
\mbox{Jia Zeng\textsuperscript{1\dag}}, \mbox{Hongrui~Zhu\textsuperscript{1}}
\par
}

\noindent 

\sloppy
\subsection*{\color{cvprblue}\textbf{Contributors}}
\mbox{Zetao~Cai\textsuperscript{1}},
\mbox{Jiafei~Cao\textsuperscript{1}}, \mbox{Yilun~Chen\textsuperscript{1}}, \mbox{Zeyu~He\textsuperscript{1}}, \mbox{Lei~Jiang\textsuperscript{2}}, \mbox{Hang~Li\textsuperscript{1}}, \mbox{Hengjie~Li\textsuperscript{1}}, \mbox{Yufei~Liu\textsuperscript{2}}, \mbox{Yanan~Lu\textsuperscript{1}}, \mbox{Qi~Lv\textsuperscript{1}}, \mbox{Yu~Qiao\textsuperscript{1}}, \mbox{Yanqing~Shen\textsuperscript{1}}, 
\mbox{Xu~Shi\textsuperscript{1}}, \mbox{Bolun~Wang\textsuperscript{1}}, \mbox{Hanqing~Wang\textsuperscript{1}}, \mbox{Jiaheng~Wang\textsuperscript{1}},
\mbox{Tai~Wang\textsuperscript{1}}, \mbox{Xueyuan~Wei\textsuperscript{1}}, \mbox{Chao~Wu\textsuperscript{1}}, 
\mbox{Yiman~Xie\textsuperscript{1}}, \mbox{Boyang~Xing\textsuperscript{2}}, \mbox{Yuqiang~Yang\textsuperscript{1}}, \mbox{Yuyin~Yang\textsuperscript{1}}, \mbox{Qiaojun~Yu\textsuperscript{1}}, \mbox{Feng~Yuan\textsuperscript{1}}, 
\mbox{Jingjing~Zhang\textsuperscript{1}}, \mbox{Shenghan~Zhang\textsuperscript{1}},
\mbox{Shi~Zhang\textsuperscript{1}}, \mbox{Zhuoma~Zhaxi\textsuperscript{1}}, \mbox{Bowen~Zhou\textsuperscript{1}}, \mbox{Yuanzhen~Zhou\textsuperscript{1}}, \mbox{Yunsong~Zhou\textsuperscript{1}},  \mbox{Yangkun~Zhu\textsuperscript{1}}, \mbox{Yuchen~Zhu\textsuperscript{2}}

\footnotetext[1]{Shanghai AI Laboratory \quad \textsuperscript{2}Humanoid Robot (Shanghai) Co., Ltd. \quad \textsuperscript{$\dag$}Project lead \quad
\textsuperscript{$\natural$}Corresponding author}

% \section{Ackownledgement}
% \input{sections/appendix_interface}
% \input{sections/appendix_rich_study}

\end{document}